\pdfoutput=1
\documentclass{article}
\usepackage{spconf,amsmath,epsfig, graphicx, subfigure}
\usepackage{float, amsfonts, caption, mathrsfs, multirow, diagbox}
\title{CYCLONE INTENSITY ESTIMATE WITH CONTEXT-AWARE CYCLEGAN}
\name{Yajing Xu, Haitao Yang, Mingfei Cheng, Si Li}
\address{School of Information and Communication Engineering,\\Beijing University of Post and Telecommunications, Beijing, China}
\begin{document}
\maketitle
\begin{abstract}
Deep learning approaches to cyclone intensity estimation have recently shown promising results. However, suffering from the extreme scarcity of cyclone data on specific intensity, most existing deep learning methods fail to achieve satisfactory performance on cyclone intensity estimation, especially on classes with few instances. To avoid the degradation of recognition performance caused by scarce samples, we propose a context-aware CycleGAN which learns the latent evolution features from adjacent cyclone intensity and synthesizes CNN features of classes lacking samples from unpaired source classes. Specifically, our approach synthesizes features conditioned on the learned evolution features, while the extra information is not required. Experimental results of several evaluation methods show the effectiveness of our approach, even can predicting unseen classes.
\end{abstract}
\section{Introduction}
\label{sec:intro}
Cyclone intensity estimation is an important task in meteorology for predicting disruptive of the cyclone. The intensity of a cyclone, which is defined as the maximum wind speed near the cyclone center, is the most critical parameter of a cyclone \cite{cyclone, CycloneClassify}. The main assumption of estimation method is that cyclones with similar intensity tend to have a similar pattern \cite{cyclone}. The cyclone features represented by the early estimation approaches rely on the human-constructed features which are sparse and subjectively biased \cite{cyclone}. With remarkable progress of deep learning, using Convolutional Neural Networks (CNN) to estimate the intensity of cyclones \cite{CycloneClassify, CycloneRotation} has obtained increasing attentions.

\begin{figure}[h]
\centering
\subfigure[]{
\begin{minipage}[b]{0.14\columnwidth}
\includegraphics[width=1.3cm, height=1.3cm]{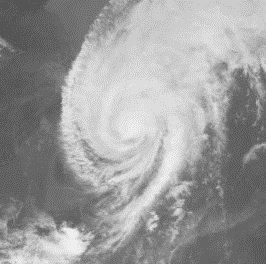}\vspace{0pt}
\includegraphics[width=1.3cm, height=1.3cm]{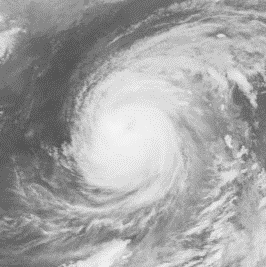}\vspace{-5pt}
\end{minipage}
\begin{minipage}[b]{0.14\columnwidth}
\includegraphics[width=1.3cm, height=1.3cm]{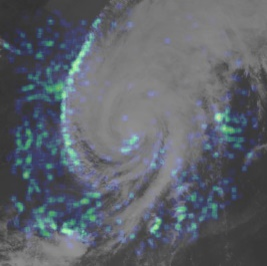}\vspace{0pt}
\includegraphics[width=1.3cm, height=1.3cm]{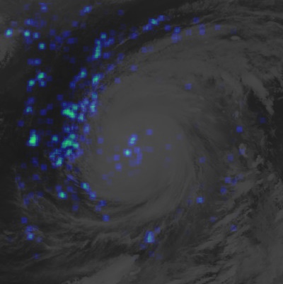}\vspace{-5pt}
\end{minipage}
\begin{minipage}[b]{0.14\columnwidth}
\includegraphics[width=1.3cm, height=1.3cm]{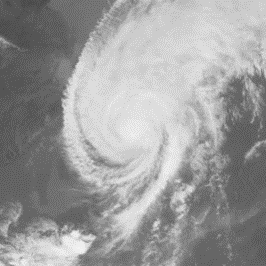}\vspace{0pt}
\includegraphics[width=1.3cm, height=1.3cm]{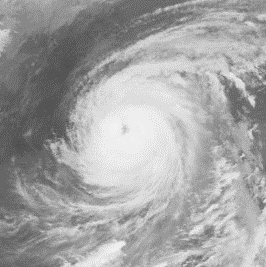}\vspace{-5pt}
\end{minipage}
\hspace*{0.08cm}%
\begin{minipage}[b]{0.14\columnwidth}
\includegraphics[width=1.3cm, height=1.3cm]{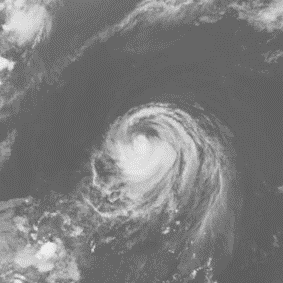}\vspace{0pt}
\includegraphics[width=1.3cm, height=1.3cm]{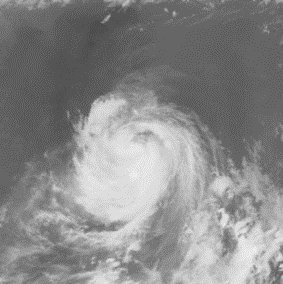}\vspace{-5pt}
\end{minipage}
\begin{minipage}[b]{0.14\columnwidth}
\includegraphics[width=1.3cm, height=1.3cm]{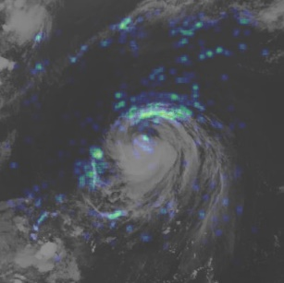}\vspace{0pt}
\includegraphics[width=1.3cm, height=1.3cm]{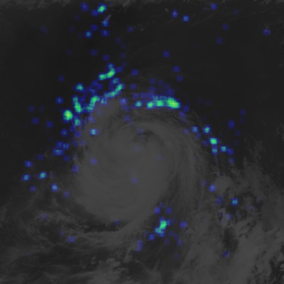}\vspace{-5pt}
\end{minipage}
\begin{minipage}[b]{0.14\columnwidth}
\includegraphics[width=1.3cm, height=1.3cm]{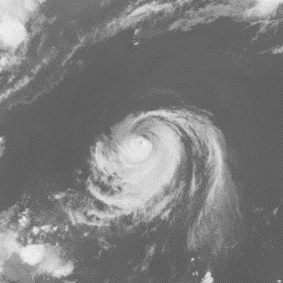}\vspace{0pt}
\includegraphics[width=1.3cm, height=1.3cm]{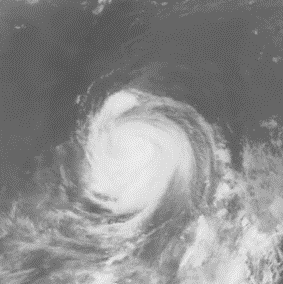}\vspace{-5pt}
\end{minipage}
\vspace{-8pt}
} 
\subfigure[]{
\begin{minipage}[b]{1\columnwidth}
\includegraphics[width=1\columnwidth,height=4.3cm]{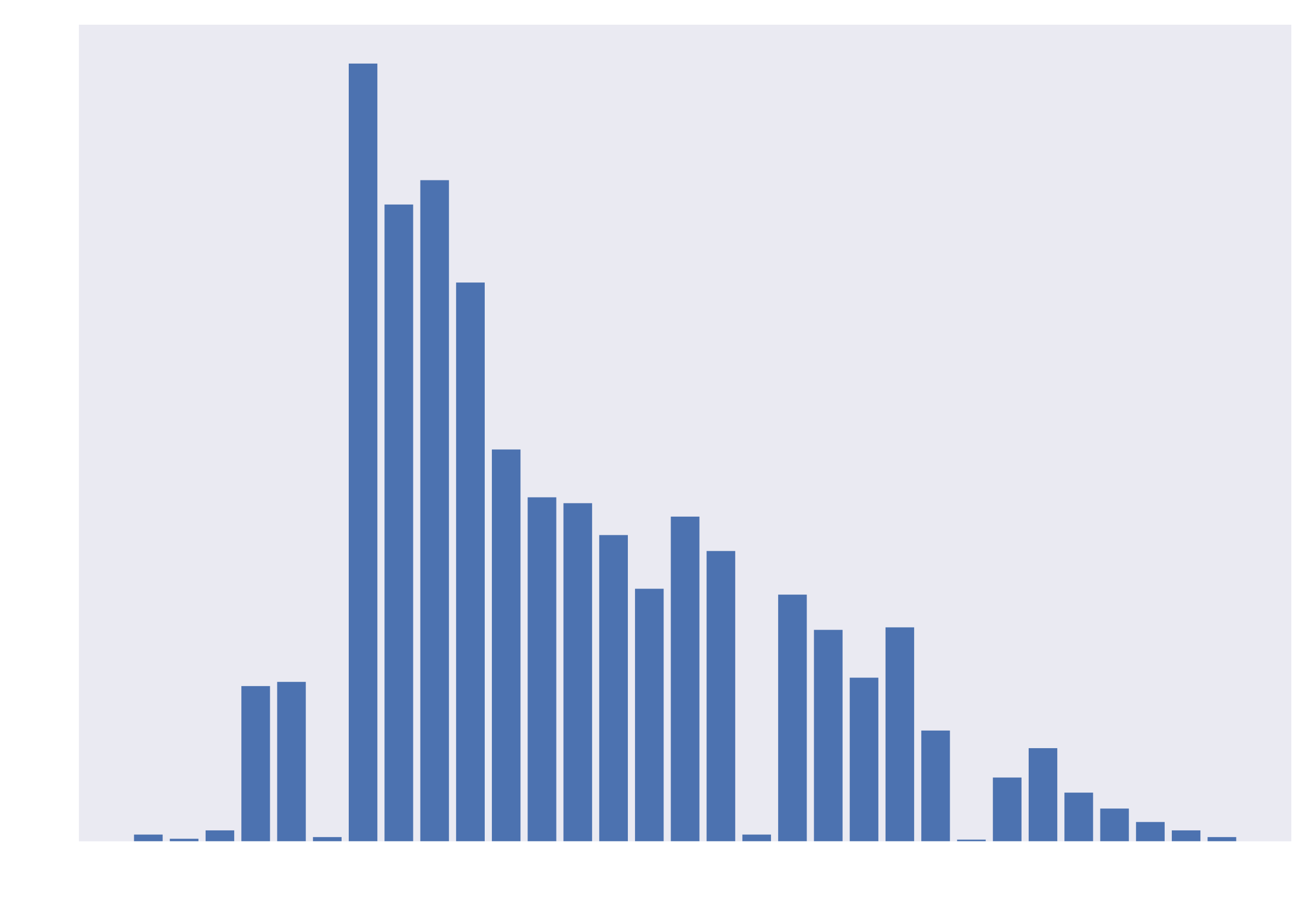}\vspace{-8pt}
\end{minipage}
}
\caption{ (a) shows four groups of cyclone images. In each group, the highlighted points of the middle image denote evolution features between the left image and the right image. (b) is the statistical results of cyclone images with different intensities in datasets.}\vspace{-5pt}
\label{fig:example}
\end{figure}
Existing deep learning approaches for cyclone intensity estimation can be roughly split into two categories: classification based methods \cite{CycloneClassify} and regression based methods \cite{CycloneRotation}. Classification approaches estimate cyclone intensities by treating each intensity label as an independent fixed class and use a cross-entropy loss to optimize model. Regression methods estimate exact intensity values of cyclones by using mean squared error (MSE) as a loss function. However, a common intrinsic problem in existing approaches is that they all ignore negative effects of specific cyclone data distribution, as shown in Fig 1 (b), where some cyclone classes only contain few instances.\par
\begin{figure*}[h]
\centering
\setlength{\abovecaptionskip}{0.cm}
\setlength{\belowcaptionskip}{-0.cm}
\includegraphics[width=0.8\textwidth, height=4.6cm]{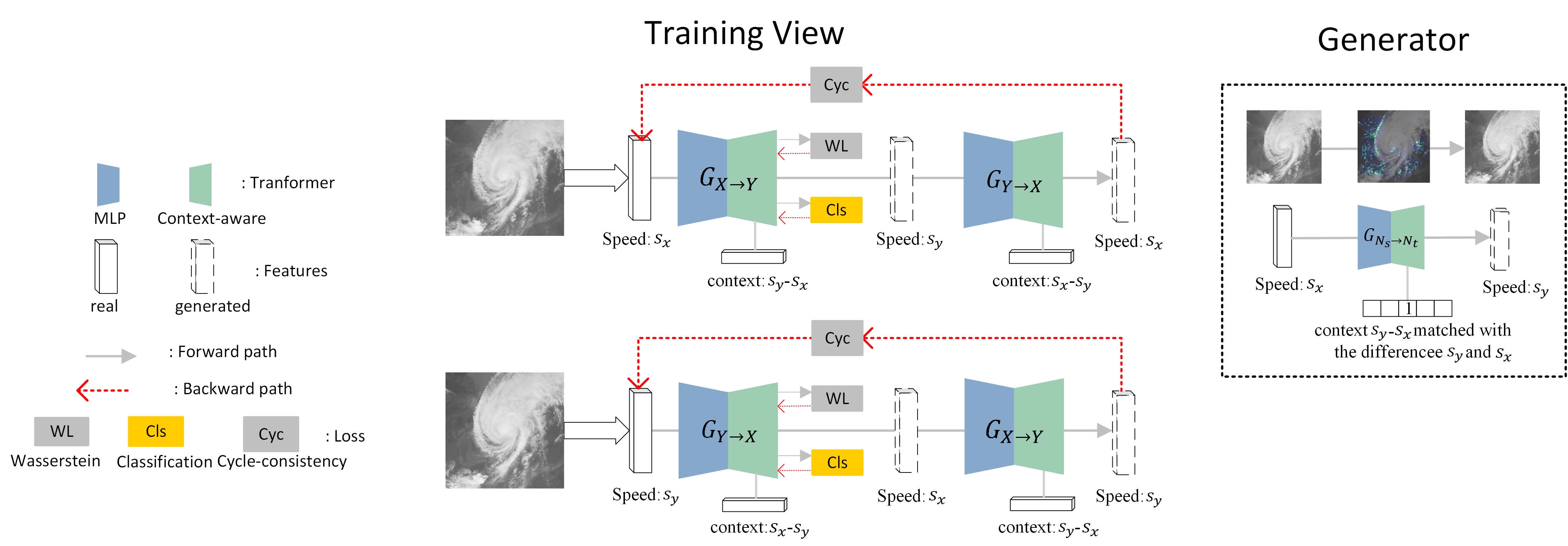}
\caption{The illustration of our context-aware CycleGAN consists of a training view (left) and a Generator (right). \textbf{Training View} shows that two generators learn the evolution feature between X and Y. These tow generator are supervised by the adversarial loss and classification loss (supported by pre-trained classifier) with speed label and context information. Discriminator $D_X$ and $D_Y$ are not shown for simplify. \textbf{Generator} shows the details of generator in our context-aware CycleGAN}\vspace{-15pt}
\label{fig:pipline}
\end{figure*}
A natural approach to address this problem is synthesizing required samples to supplement training set. Recently, generating training data with generative adversarial networks (GAN) obtains increasing attentions due to the ability of generation conditioned on specific attributes or categories, e.g. synthesize disgust or sad emotion images from neutral class with Cycle-Consistent Adversarial Networks (CycleGAN) \cite{Emotion}. Most of existing approaches \cite{featureGeneration, featureGenerationCycle} rely on extra attribute information to train GAN model and generate samples. However, in this work, each cyclone sample only has an intensity label and some cyclone classes are extremely lacking samples, which make existing methods perform poorly. Evolution features, the difference of features between adjacent cyclone classes, has a similar pattern on each fixed cyclone intensity interval. As illustrated in Fig.1 (a), the highlighted points are evolution features between images of adjacent cyclone intensities, which are located on the eye of cyclones and the border of cyclones and sea.  

In this paper, we propose a context-aware CycleGAN to synthesize CNN features of cyclone classes lacking samples in the absence of extra information. In particular, the generator of context-aware CycleGAN is modified to learn cyclone evolution features conditioned on a given context intensity interval which is the difference in intensity between each two samples involved in training. Moreover, the generator is regularized by a classification loss to regulate the intensity characteristic of synthetic features and the adversarial loss is improved with the Wasserstein distance \cite{WGAN} for a training stability. Based on the learned evolution features, our method is able to synthesize features of any cyclone intensity with a contextual cyclone sample.

We summarize our contributions as follows:
\begin{itemize}
\item Propose the concept of evolution features which focus on the difference of features between adjacent classes instead of normal features of classes.
\item Improve CycleGAN to synthesize features under the constraint of a context intensity interval where a classification loss is optimized to regulate cyclone intensity.
\end{itemize}

\section{Our approach}
\label{sec:format}

Different from conventional generative methods, our method is suitable to synthesize features for classes lacking samples and classes which have context-dependent without extra information. Specific to the cyclone intensity estimation, the context refers to the interval of adjacent cyclone intensities. Therefore, the key of our model is the ability to generate required CNN features from unpaired source classes by relying on evolution features of a fixed context intensity interval shown in Fig 2.

We begin by defining the problem of our interest. Let $X={(f_x,s_x,c_{x\rightarrow y})}$ and $Y={(f_y,s_y,c_{y\rightarrow x})}$ where X and Y are source and target classes, $f\in \mathbb{R}^{d_x}$ is the CNN features of cyclone image extracted by the CNN, $s$ denotes the intensity labels in $S=\{s_1,...,s_K\}$ consisting of K discrete speeds, $c_{x\rightarrow y}$ is the context attribute vector of label from $s_x$ to $s_y$, so as the $c_{y\rightarrow x}$. To improve the generalization ability of our method, the CNN features $f$ is connected with the random noise, as in \cite{featureGeneration, featureGenerationCycle}. 
\subsection{Context-aware CycleGAN}
\label{sec:generator}
In general, GAN \cite{WGAN} consists of a generative network ($G$) and a discriminative network ($D$) that are iteratively trained in a two-player minimax game manner. CycleGAN is used to realize the unpaired translation between source classes ($X$) and target classes ($Y$). There are two generator $G_{X\rightarrow Y}$ and $G_{Y \rightarrow X}$ in CycleGAN. $G_{X\rightarrow Y}$ learns a mapping $X\rightarrow Y$ and $G_{Y \rightarrow X}$ learns a mapping $Y \rightarrow X$ simultaneously \cite{cycleGAN}.

\noindent
\textbf{Context-aware Transform.}
Learning evolution features from adjacent cyclone classes is critical to our method. Hence, our generator consists of a single hidden layer and a context-aware transform layer conditioned on the context $c$, which focus on the evolution features.

Formally, the context-aware transform layer in the generator transforms the input $f_i$  to the output $f_o$ by relying on the context vector $c$. The transformation is denoted as $f_o=g(f,c)=r(W^c f+b)$, where the $g()$ is the function of transform layer, $r()$ is the relu function, $b$ is the bias of the layer and $W^c$ is the weight parameter. In particular, the weight parameter $W^c$ is designed as the summation of the following two terms:
\begin{equation}
W^c=\overline{W}_p+V_pE(c),
\label{weight}
\end{equation}
where $\overline{W}_p$ is the independent weight of the context attributes, $E(c)$ is the desired shape context representation, which is turned from the context intensity interval by using a function denoted as $E(·)$, and $V_p$ transforms the context attributes to the weight of auxiliary transform layer. All of parameters in \eqref{weight} are learnt during the training stage \cite{contextTranfer, context-aware}. Moreover, the second term in \eqref{weight} is an auxiliary transform layer generated from the context attribute $c$, which focuses on learning evolution features. 

\noindent
\textbf{Objective Function.}
Only using evolution features do not guarantee label-preserving transformation. For generating features of expected intensity, the classification loss of pertained classifier is proposed to be minimized as the regular term of generator. Besides, adversarial losses in CycleGAN are applied to iteratively train the generators and discriminators \cite{cyclone}.

In particular, the regularized classification loss $L_{cls}$ is defined as:
\begin{equation}
\mathcal {L}_{cls}(s_y, \widetilde{f}_y;\theta)= -E_{\widetilde{f}_y\sim P_{\widetilde{f}_y}}[log P(s_y|\widetilde{f}_y;\theta)],
\vspace{-8pt}
\label{classification_loss}
\end{equation}
where $\widetilde{f}_y$ is features synthesized by $G_{X\rightarrow Y}$ from the real features $f_x$, $s_y$ is the class label of $\widetilde{f}_y$ and $(s_y|\widetilde{f}_y;\theta)$ denotes the probability of the synthetic features predicted with its true label $s_y$ which is the output of liner softmax with the parameter $\theta$ pretrained on real features. $L_{cls}(s_x, \widetilde{f}_x;\theta)$ is analogously defined with the same parameter $\theta$. Hence, the full objective is:
\vspace{-3pt}
\begin{equation}
\begin{aligned}
\mathcal{L}(G_{X\rightarrow Y}, &G_{Y\rightarrow X}, D_X, D_Y) =\mathcal{L}_{GAN}(G_{X\rightarrow Y},D_Y)\\+
&\mathcal{L}_{GAN}(G_{Y\rightarrow X},D_X)+
\lambda_1 \mathcal{L}_{cyc}(G_{X\rightarrow Y}, G_{Y\rightarrow X})\\+
&\beta(\mathcal{L}_{cls}(s_y, \widetilde{f}_y;\theta)+\mathcal {L}_{cls}(s_x, \widetilde{f}_x;\theta)),
\vspace{-8pt}
\label{all_loss}
\end{aligned}
\end{equation}
where $\beta$ is a balance term to weight the classification loss, The first three terms are adversarial loss. 

In order to improve the stability of training, the adversarial loss $L_{GAN}(G_{X\rightarrow Y}, D_Y)$ and  $L_{GAN}(G_{Y\rightarrow X},D_X)$ are the Wasserstein loss of WGAN \cite{WGAN}. Besides, the generators rely on the context attributes which can be considered as a special situation of conditional GAN \cite{conditionalGAN, contionalCycleGAN}. Therefore, the context intensity interval is also fed into the discriminator. The conditional adversarial loss with Wasserstein distance \cite{WGAN, WGAN-GP} are defined as: 
\begin{equation}
\begin{scriptsize}
\begin{aligned}
\mathcal {L}_{GAN}(G_{X\rightarrow Y}, &D_Y)=E[D_Y(f_y,c_{x\rightarrow y})]- E[D_Y(\widetilde{f}_y, c_{x\rightarrow y})]\\ -&\lambda_2 E[(\|{\nabla}_{\widehat{f}_{y}}D_Y(\hat{f}_{y}, c_{x\rightarrow y})\|_2-1)^2],
\end{aligned}
\end{scriptsize}
\label{generate loss}
\end{equation}
\noindent
where $\widetilde{f}_y=G_{X\rightarrow Y}(f_x, c_{x\rightarrow y})$, $\widehat{f}_{y}=\alpha f_y+(1-\alpha)$ with $\alpha \sim U(0, 1)$ and $\lambda_2$ is the penalty coefficient. The first two terms approximate the Wasserstein distance and the third term is the gradient penalty\cite{WGAN-GP}. And $L_{GAN}(G_{Y\rightarrow X}, D_X)$ is defined analogously as equation (4).

\begin{scriptsize}
\end{scriptsize}
\subsection{Synthesizing\&Classification}
\label{sec:classification}
Based on the structure and loss function described above, the generator is trained to learn the evolution features of adjacent cyclone classes. With a real specific cyclone sample as input, the trained context-aware CycleGAN relies on the evolution features which are learned from the fixed context interval to synthesize CNN features of the cyclone whose intensity is adjacent to the input sample. Moreover, generated features is allowed to train a classifier with real features. The standard softmax classifier parameter $\theta_{cls}^*$ are minimized with the cross-entropy loss and the final prediction is defined as follows:
\begin{table}[htb]
\small
\begin{center}
\begin{tabular}{c|c|c|c}
\hline
Methods&f1(\%)&MAE&RMSE\\
\hline
Pradhan e.t(2018)\cite{CycloneClassify}*&56.63&3.05&6.21\\
Chen e.t(2018)\cite{CycloneRotation}*&51.27&4.27&8.30\\
AlexNet &60.30&2.61&5.94\\
ResNet &57.58&\textbf{2.37}&\textbf{4.97}\\
Ours&\textbf{61.41}&2.49&5.70\\
\hline
\end{tabular}
\caption{The estimation result of different methods. * denotes the implement result with the original model.} \label{tab:CNN}\vspace{-10pt}
\end{center}
\end{table}
\vspace{-3pt}
\begin{equation}
y^*= argmax P(y|f_x, \theta_{cls}^*),
\vspace{-3pt}
\end{equation}
where $f_x$ is features of a testing sample and $y^*$ is the final predicting intensity.

\section{Experiments}
\label{sec:pagestyle}
In this section, we firstly introduce the details of our implementation, then the experimental details including datasets, evaluation criteria and features used in our experiments are introduced. Finally, results of our approach compared with other estimation models especially on classes with few training instances are shown.
\subsection{Implementation Details}
Both of the discriminator $D_X$ and $D_Y$ consist of two MLP layers with LeakyReLU activation and the hidden layers of discriminator contain 4096 hidden units due to the application of Wasserstein loss. We use Adam to optimize the classifier and SGD to optimize our context-aware CycleGAN. And the learning rate is 0.0001 with decay 0.9 for every 10 epochs while training the classifier and 0.001 while training the basic CNN network. The balance term $\beta$ in the loss function is 0.001 and the hyperparameters are $\lambda_1$=10 and $\lambda_2$=10 as suggested in \cite{WGAN-GP, cycleGAN}. All noises are independent and drawn from a unit Gaussian distribution with 128 dims. We also do not apply batch normalization as empirical evaluation showed a significant degradation of the result when batch normalization is used.
\subsection{Experiment Details}
\textbf{Datasets and Evaluation Criteria}. The total 7686 cyclone images with 151 cyclone intensities are collected from Pacific Ocean. And we train our model by using 6118 training images and perform estimation results on 1560 testing images. The cyclone image example is shown in Fig 1 and the number of images with different intensities is in Fig 1 (b).  Besides, the size of cyclone image is $400\times400$. Note that the center of cyclones are all placed at the middle grid of each image and data augmentation is abbreviated as “Data Aug” in following experiments. 
\begin{table*}[htb]
\centering
\begin{scriptsize}
\resizebox{\textwidth}{!}{
\begin{tabular}{|c|cccccc|cccccc|cccccc|}
\hline
\multirow{2}{*}{Methods}&\multicolumn{6}{c}{f1-score(\%)}& \multicolumn{6}{c}{MAE}&\multicolumn{6}{c|}{RMSE}\\
\cline{2-19}
&14&16&17&43&55&65&14&16&17&43&55&65&14&16&17&43&55&65 \\
\hline
 chen(2018)&0&36.36&0&0&46.42&42.85&14.6&4.86&9.0&8.12&5.61&17.57&18.86&8.96&13.30&11.06&10.65&24.86\\
 AlexNet&0&44.73&0&0&52.00&57.14&14.5&3.36&3.60&6.75&2.76&8.85&18.83&7.61&4.60&8.23&6.47&16.53\\
 Data Aug&0&\textbf{50.50}&0&0&46.66&55.55&12.83&2.79&3.80&6.75&2.23&\textbf{6.42}&18.09&7.35&4.66&8.23&4.73&\textbf{15.23}\\
 Ours&\textbf{26.66}&50.00&\textbf{25.00}&\textbf{50.00}&\textbf{54.54}&\textbf{66.66}&\textbf{8.50}&\textbf{2.29}&\textbf{3.40}&\textbf{4.62}&\textbf{2.09}&\textbf{6.42}&\textbf{14.70}&\textbf{4.46}&\textbf{4.58}&\textbf{7.37}&\textbf{4.60}&\textbf{15.23}\\
 \hline
\end{tabular}
}
\end{scriptsize}
\caption{The result of other method and our proposed context-aware CycleGAN on intensity classes lacking training instances. The best results are highlighted. We do rotations for these classes training instance denote as the “Data Augmentation”. Note that not all cyclone classes are shown as we focus on those with few training instances and cyclone  intensity is  measured in m/s.}\vspace{-20pt}
\label{tab:instances}
\end{table*}

Following the evaluation metrics in \cite{CycloneClassify}, we use f1-score, mean absolute error (MAE) and root-mean-square intensity error (RMSE). The estimated intensities are determined as the category with the highest probability. MAE and RMSE measure the error between estimated intensities and actual intensities, which present the error in the form of average absolute value or square of two intensity difference in m/s.  The f1-score is widely used to measure the performance of classification models.

\noindent
\textbf{Features}. To extract effective real CNN features, we compare the performance of ResNet \cite{ResNet}, AlexNet \cite{AlexNet} and other CNN structures in \cite{CycloneRotation, CycloneClassify}, which are the latest related approaches in the cyclone intensity estimation task. It is worth nothing that the dimension of the fully connected layers in above structures are adjusted to accommodate the image size and desired feature dimension. And the performance of these models is shown in Table \ref{tab:CNN}. We can find that the performance of AlexNet is the best. Therefore, we use the 512-dim visual feature extracted by the AlexNet as the representation of the cyclone images. Besides, we generate 512-dim CNN features with our context-aware CycleGAN for synthetic features of cyclone images.
\begin{figure}[htbp]
\begin{center}
\includegraphics[width=1\columnwidth]{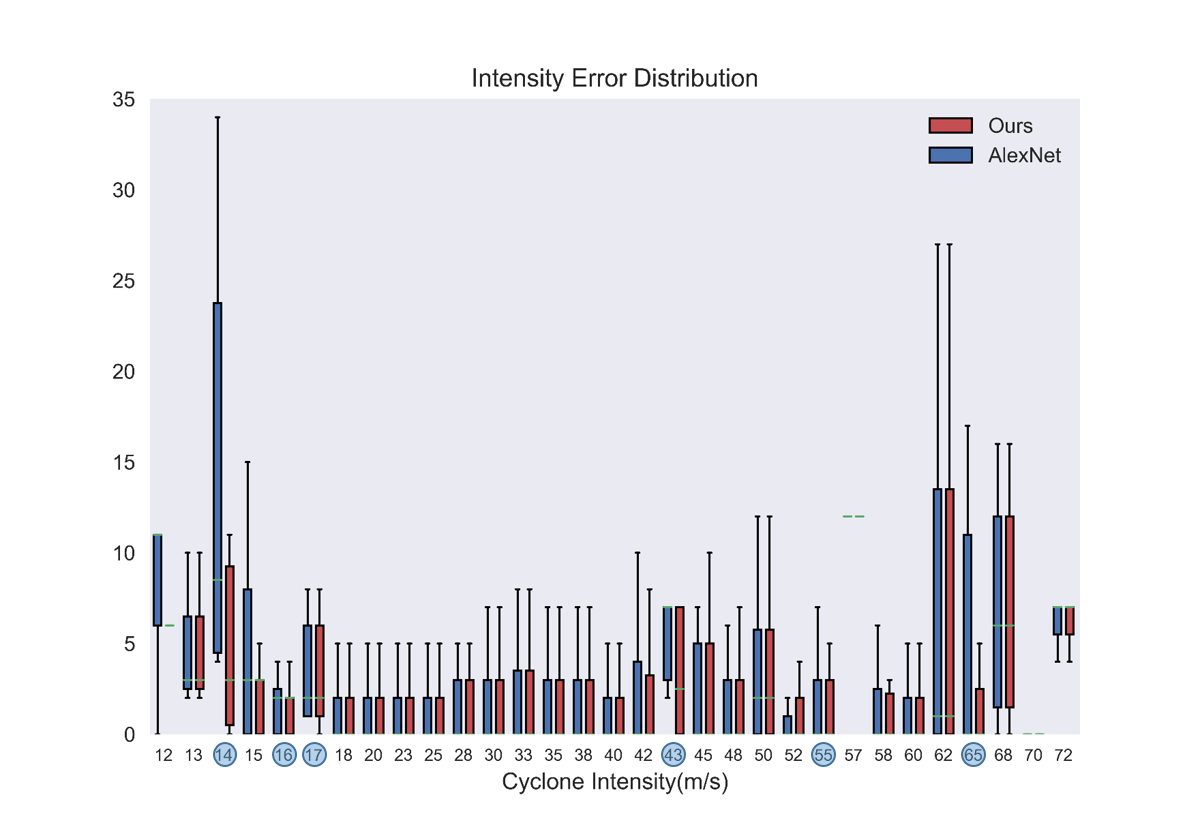}
\caption{The distribution of the absolute error between estimate and truth cyclone intensity with same setting as Table \ref{tab:instances}. The highlighted labels indicate that the generated features of these classes are mixed with the real features to train the classifier.}\vspace{-20pt}
\label{fig:pipline}
\end{center}
\end{figure}
\subsection{Results and Analysis}

\noindent
\textbf{Effect of CNN Architectures.} Due to the CNN encoder provides real features which is the important guiding information to the discriminator, the effect of CNN encoder is critical in our experiments. Hence, we compare the performance of different CNN encoders and estimation results are shown in Table \ref{tab:CNN}. The prominent result is that ResNet with a deeper architecture has worse performance than other models on f1-score. We conjecture that this problem may be caused by the quality and size of training dataset, e.g. the noise of images produced by the sea fogs. In addition, the size of dataset is a major concern in our experiments, which may cause overfitting of CNN encoders. Moreover, in our experiments, the model in Chen et al.\cite{CycloneClassify} which is designed for the regression task dose not perform better than the model in Pradhan et al.\cite{CycloneRotation} designed for classification.

\noindent
\textbf{Classification with Generated Features.} The experimental results in Table \ref{tab:instances} clearly demonstrate the advantage of our proposed context-aware CycleGAN in cyclone intensity estimation. It is obviously that using features of cyclone classes lacking samples generated by context-aware CycleGAN to train the classifier provides remarkable improvements over the basic models represented by AlexNet. In particular, our approach achieves the best estimation results in 5 out of 6 the cyclone classes on three evaluation metrics. The only exception is f1-score on intensity 16, where the best result is achieved by Data Augmentation. Besides, the Data Augmentation achieves the same MAE and RMSE results on intensity 65 as our context-aware CycleGAN. Compared to Data Augmentation, our approach achieves a better f1-score with same MAE and RMSE on intensity 60 and a lower error on intensity 16, which prove the effectiveness of our approach. A critical question of cyclone intensity estimation is the error distribution, especially for the classes lacking samples. To address this question, we show the error distribution of estimation and truth cyclone intensity in Fig 3. The significantly smaller rectangles prove the effectiveness of our method.
\section{conclusion}
\label{sec:conclusion}

In this paper, we propose context-aware CycleGAN to generate features depended on learned evolution features of a context interval without extra information. Our proposed aprraoch is able to tackle the problem related to extreme scarcity of data while context evolution features are existing between different classes, e.g. cyclone intensity estimation in this paper. Experiments on the cyclone dataset have shown significant improvements with synthetic features on classes lacking samples. Further, we will investigate more effective representation for cyclone images.

\section{acknowledge}
\label{sec:acknowledge}
This work was supported by National Natural Science Foundation of China (61702047), Beijing Natural Science Foundation (4174098) and the Fundamental Research Funds for the Central Universities (2017RC02).

\bibliographystyle{IEEEbib}
\bibliography{arixv_cyclone.bib}
\end{document}